\documentclass{llncs}
\pdfoutput=1
\usepackage{makeidx}  
\usepackage{algorithmic}
\usepackage{algorithm}
\usepackage{graphicx}
\usepackage{url}
\usepackage{hyperref}
\usepackage{amssymb,amsmath}
\newcommand{\vect}[1]{\mathbf{#1}}
\usepackage{bm}

\usepackage{multirow}

\DeclareMathOperator*{\argmin}{arg\,min}
\begin{document}
\frontmatter          
\pagestyle{headings}  

\mainmatter              
\title{Collision-Free Poisson Motion Planning in Ultra High-Dimensional Molecular Conformation Spaces}
\titlerunning{Collision-free Poisson planning}
\author{Rasmus Fonseca\inst{1,2} \and Dominik Budday\inst{3}
 \and Henry van den Bedem\inst{2}}
\authorrunning{Fonseca et al.} 
\institute{Stanford University, Molecular and Cellular Physiology, Stanford, CA, USA\\
\email{rfon@stanford.edu},
\and
SLAC National Accelerator Laboratory,
Bioscience Division, Stanford University\\ Menlo Park, CA, USA\\
\email{vdbedem@stanford.edu}
\and
{University of Erlangen-Nuremberg, Chair of Applied Dynamics\\ 91058 Erlangen, Germany}}

\maketitle              

\begin{abstract}
The function of protein, RNA, and DNA is modulated by fast, dynamic exchanges between three-dimensional conformations. Conformational sampling of biomolecules with exact and nullspace inverse kinematics, using rotatable bonds as revolute joints and non-covalent interactions as holonomic constraints, can accurately characterize these native ensembles. However, sampling biomolecules remains challenging owing to their ultra-high dimensional configuration spaces, and the requirement to avoid (self-) collisions, which results in low acceptance rates.
Here, we present two novel mechanisms to overcome these limitations. First, we introduced temporary constraints between near-colliding links. The resulting constraint varieties instantaneously redirect the search for collision-free conformations, and couple motions between distant parts of the linkage. Second, we adapted a randomized Poisson-disk motion planner, which prevents local oversampling and widens the search, to ultra-high dimensions. We evaluated our algorithm on several model systems. Our contributions apply to general high-dimensional motion planning problems in static and dynamic environments with obstacles.
\end{abstract}
\section*{Introduction}
Proteins interact with their partners and perform their cellular functions by rapidly interchanging between three-dimensional substates~\cite{VandenBedem2015}. Computational methods to readily characterize the conformational landscape of folded proteins or their complexes would help us interpret ensemble-averaged, experimental data, and can provide insight in how they function. Molecular dynamics simulations can reveal time-resolved, atomically detailed trajectories, but require sophisticated resources to overcome spatiotemporal barriers separating functional substates~\cite{Dror2012-ai}. By contrast, non-deterministic conformational sampling procedures, such as Monte Carlo sampling, can rapidly collect a representative set of conformations but disregard time-scales of molecular change. 

Fast, robotics-inspired algorithms for randomized exploration of molecular conformational spaces are quickly emerging as an alternative to these more traditional molecular simulations \cite{Al-Bluwi2012b}. They are frequently designed around geometric motion planners, such as probabilistic roadmaps (PRMs, \cite{kavraki1996probabilistic,thomas2007simulating}), rapidly-exploring random trees (RRTs, \cite{lavalle2000rapidly,cortes2005path,kirillova2008nma}) or their variants \cite{Noe2006,Yao2012}. While these planners efficiently handle high-dimensional configuration spaces, they require adaptations to sample broadly and uniformly in molecular simulations~\cite{Yao2012,cortes2005path}. 

\begin{figure}[bt]
\centering
\includegraphics[width=1.0\textwidth]{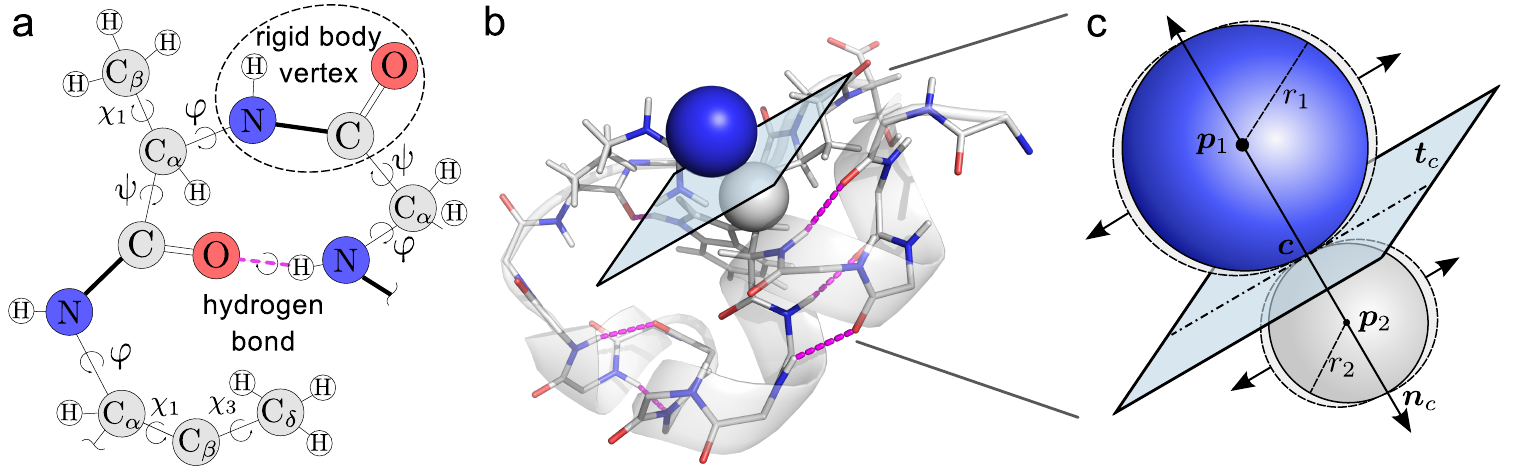}
\caption{Kinematic representation of molecules and constraints. a) A molecular graph with double bonds and partial double bonds highlighted. After edge-contractions these atoms become a single rigid body vertex. The remaining bonds can rotate around the bond axis and are described by dihedral angles. b) Small section of a protein molecule with hydrogen bonds marked as purple lines and a nitrogen-hydrogen clash marked with van der Waals spheres and the bisecting plane. c) A dynamic clash-avoiding constraint (dCC) allows individual motions of the near-clashing atoms $\bm{p}_1$ and $\bm{p}_2$ along directions $\bm{t}_c$ in the plane with normal vector $\bm{n}_c$, but only a joint move along $\bm{n}_c$. This allows atoms to slide past each other, but prevents them from getting closer.}
\label{fig:clashPanels}
\end{figure}

These approaches often represent a protein as a kinematic linkage, with groups of atoms as rigid links and rotatable bonds as revolute joints (Fig.~\ref{fig:clashPanels}.a). Randomly perturbing the rotatable bonds would quickly lead to unfolding the protein. Instead, the rotatable bonds require coordinated changes to maintain non-covalent interactions, such as hydrogen bonds, in the protein. Nullspace inverse kinematics~\cite{Burdick1989} is an efficient technique to deform a protein while observing constraints~\cite{VandenBedem2005,Yao2012}. In nullspace inverse kinematics, we express non-covalent interactions in the protein as holonomic constraints~\cite{Budday2015a}, which define a lower-dimensional variety in conformational space~\cite{Burdick1989}. Deformations of the protein structure on the constraint variety therefore maintain the folded state.

Alternative, non-native contacts such as steric interactions by repulsive van der Waals forces (clashes) also play a critical role in protein conformational dynamics by stabilizing native states and redirecting the motion of coordinated degrees-of-freedom (DOF, \cite{Oliveberg2005}). However, clashes also severely hinder fast exploration of conformational space. This is particularly the case for proteins, whose cores are densely packed. While several techniques have been developed to address steric hindrance and increase the efficiency of conformational sampling, most neglect the role of non-native contacts by avoiding, retrospectively relieving, or simply ignoring clashes.  For example, a common solution is to accept or reject a trial conformation after checking for clashes \cite{cortes2005path}. Alternatively, highly reduced structural representations of the protein \cite{haspel2010tracing,al2013modeling} or prior information constraints and filtering were proposed to avoid steric clashes \cite{Raveh2009}. Another, highly efficient, approach is to combine a reduced structural representation with normal modes as the search space for an RRT \cite{kirillova2008nma}. In the latter case, the side chains were only adjusted to relieve clashes once a new conformation was found. 

These solutions either increase the efficiency of the algorithms at the expense of atomically detailed collision avoidance, or employ expensive collision checking. However, most importantly, they fail to account for functionally important, sterically coupled motions in proteins that govern their conformational dynamics. 

Here, we address both limitations simultaneously by planning collision-avoiding motions. We capitalize on the observation that protein kinematic linkages are highly redundant. Sugiura {\em et al.}~\cite{Sugiura2007} earlier proposed velocity-based nullspace techniques to avoid self-collisions by introducing virtual, repulsive forces between near-colliding links. Petric {\em et al.}~\cite{Petric2013} recently proposed to project desired Cartesian velocities away from fixed obstacles onto nullspace joint velocities. In these solutions, collisions are locally resolved by prescribing desired velocities. 

{\bf Our Contributions.} We take a different approach that avoids explicitly prescribing velocities, but instead lets the global deformation determine how local collisions are resolved. We introduce a temporary one-dimensional constraint between near-colliding links, thereby altering the constraint variety of the linkage. The constraint allows only a joint motion along a line through their centers, or a 'sliding' motion in the plane normal to this line (Fig.~\ref{fig:clashPanels}.b,c). Velocities on this new constraint variety are collision-avoiding, corresponding to a new search direction in conformational space. By adding desired constraints, in successive sampling steps our algorithm hops between collision-free constraint varieties.

We combined our clash-avoiding procedure with a new motion planner we call \textsc{PoissonExplore} inspired by Poisson disk sampling~\cite{lagae2008comparison}. Poisson Disk sampling generates random samples that are close together, but separated by a minimum distance to cover conformation space broadly and uniformly. Recently, Manocha and coworkers introduced Poisson-RRT: a Poisson Disk-based motion planner~\cite{park2014poisson}. Despite favorable sampling properties, the complexity of Poisson planners scales exponentially with the dimension of conformation space. We designed an efficient multi-query randomized Poisson disk-based planner and adapted it to ultra-high dimensional molecular conformational spaces using a bounding volume hierarchy (BVH, \cite{angulo2005,lotan2004algorithm}) for fast neighborhood queries.

Our contributions allow us to efficiently explore the conformational landscape of biomolecules, from loop motions to large, whole molecules with hundreds or thousands of degrees of freedom. It identifies collective motions, propagated by native and non-native contacts through constraints, which give insight into the mechanisms of molecular functions.

\section*{Methods}
We designed our algorithms around the Kino-Geometric Sampling (KGS) software framework (\url{https://simtk.org/projects/kgs/}, \cite{Yao2012,Fonseca02092014}). The purpose of KGS is to efficiently explore feasible regions of a molecule's conformational space. To represent a conformation we first construct a molecular graph, $G_m=(V_m,E_m)$, such that $V_m$ contains all atoms and $E_m$ contains all covalent bonds (Fig.~\ref{fig:clashPanels}.a). Next, a rigid-body graph $G_r=(V_r,E_r)$ is constructed from $G_m$ by edge-contracting edges that correspond to double bonds or are part of pentameric rings (RNA ribose or the proline amino acid). Vertices in $V_r$ are sets of atoms that form rigid bodies and edges in $E_r$ are revolute joints. For linearly branched multi-chain molecules like proteins and RNA, this results in a set of acyclic trees where each tree represents one chain. For generally branched molecules, we can identify the minimum spanning tree and add left-out edges as constraints. Finally, we connect each tree in $G_r$ to a super-root $v_s$ via six global DOFs resulting in a single rooted tree, $T_k=(V_k,E_k)$. Forward kinematics are defined as propagation of atom coordinate transformations from $v_s\in V_k$, along the direction of edges in $E_k$. A conformation is represented as a vector of all DOFs in $T_k$: $\mathbf{q} \in \mathbb{R}^n$ where $n=|E_k|$. 

The constraints are defined by sets of atomic pairs, $C_k$, in the molecule for which the local geometry (pair distance and angles to surrounding covalent bonds) must be maintained (see Fig.\ref{fig:clashPanels}.b). For example, in KGS, hydrogen bonds with energies below -1\,kcal/mol are detected and the acceptor/donor atom-pairs are added as constraints. Only rotation around a hydrogen bond axis is permitted, which leads to five constraints per hydrogen bond. Additionally, any extraneous covalent bonds that induce cycles in $T_k$ are added to $C_k$ as well. 

A conformation is considered feasible if covalent bond and constraint lengths as well as angles remain constant and if no atoms are clashing. The geometry of covalent bonds is maintained implicitly by using revolute joints for the internal edges. Clashing atoms are detected by hashing atom coordinates into a 3D grid of $1\times 1 \times 1$ (\AA{}$^3$) cells and for each atom performing an expected constant-time query~\cite{halperin1994} in neighboring cells. As protein cores are extremely tightly packed we multiply their van der Waals radii with 0.75 before checking if they overlap. Taking steps in conformational space while maintaining the geometry of constraints is the responsibility of the \emph{conformational perturbations} and coordinating the use of these operators is the responsibility of the \emph{planners} as described in the following sections.

\subsection*{Conformational Perturbations}
To comprehensively explore the conformational space we employ two distinct operators termed analytical inverse kinematics (AIK) and nullspace inverse kinematics (NIK) perturbation.

The AIK perturbation introduced by Coutsias \emph{et al}.~\cite{coutsias2006} takes a sub-chain spanned by three \emph{pivot atoms} with a total of six adjacent rotational DOFs and analytically computes all possible closed conformations of the sub-chain. It is a requirement that each pivot atom is the end-point of exactly two distinct rotational axes and that there are no hydrogen bonds or other constraints in the sub-chains interior. This perturbation allows jumps between unconnected regions of the feasible conformational manifold and serves to generate distinct seed conformations for the two other more "local" perturbations.

The $5m$ holonomic constraints $\mathbf{\Phi} = \mathbf{\Phi}(\mathbf{q})$ from $m$ hydrogen bonds define a constraint variety 
\begin{equation}
\mathcal{V}_{\mathrm{hb}}=\{\mathbf{q}\in\mathbb{R}^n\;|\;\mathbf{\Phi}(\mathbf{q})=\bm{0}\}.
\end{equation}
of dimension at least $n-5m$. NIK perturbations are local perturbations on $\mathcal{V}_{\mathrm{hb}}$, which are taken from the tangent space $\mathcal{T}_\mathbf{q}(\mathcal{V}_{\mathrm{hb}})$ to $\mathcal{V}_{\mathrm{hb}}$ at $\mathbf{q}$. NIK perturbations maintain any specified constraints in linear approximation. This can be achieved by projecting a trial perturbation $\bm{\delta}_q$ onto the nullspace of the constraint Jacobian $\mathbf{J}$ evaluated at a seed conformation $\mathbf{q}$~\cite{Budday2015a}. Right-singular vectors corresponding to vanishing singular values in a singular value decomposition (SVD, \cite{golub2012matrix}) form an orthonormal basis $\mathbf{N}$ for the nullspace of $\mathbf{J}$. We obtain an admissible perturbation $\bm{\Delta}_{q}$ through projection via 

\begin{equation}
\bm{\Delta}_{q}=\vect{N}\vect{N}^T\bm{\delta}_{q}.
\label{eq:projection}
\end{equation}

In addition to constraints defined by native contacts, such as hydrogen bonds, we use a novel procedure to modulate non-native contacts by adding 1-D dynamic clash-avoiding constraints (dCC, Budday \emph{et al}.~\cite{budday2016frustration}). Whenever a perturbation in $\mathcal{V}$ leads to a prohibitive steric clash between atoms, the conformation can not be accepted. However, instead of discarding the intended search direction $\bm{\delta}_\mathbf{q}$ to find a new conformation, we redirect the perturbation onto a new variety $\mathcal{V}_\textrm{dCC}$ by introducing $c$ new dCCs that prevents each of the $c$ clashing atom-pairs from approaching each other~\cite{budday2016frustration}. Given two clashing atoms centered at $\mathbf{p}_i$ and $\mathbf{p}_j$, the dCC
\begin{equation}
\mathbf{n}_c^T\left(\frac{\partial \mathbf{p}_j}{\partial \mathbf{q}}- \frac{\partial \mathbf{p}_i}{\partial \mathbf{q}}\right){\bm{\delta}_\mathbf{q}} = 0
\label{eq:equality}
\end{equation}
on the desired perturbation $\bm{\delta}_\mathbf{q}$ allows free motions of $\mathbf{p}_1$ and $\mathbf{p}_2$ within the contact plane (Fig.~\ref{fig:clashPanels}.c)), but only a joint motion in the clash direction $\mathbf{n}_c$. In essence, the two atoms can slide past each other but maintain their distance with respect to $\mathbf{n}_c$. The constraints are formulated individually for each pair of clashing atoms and added as an additional row to the constraint Jacobian matrix $\mathbf{J}$. An SVD leads to a basis $\mathbf{N}_{\mathrm{dCC}} \in \mathbb{R}^{d\times(d-r')}$ for the equal or lower-dimensional nullspace of the Jacobian with rank $r' \geq r$. Accordingly, we obtain the corresponding, clash-avoiding nullspace perturbation using the previously introduced projection $\mathbf{\Delta}_\mathbf{q}=\mathbf{N}_{\mathrm{dCC}}\mathbf{N}_{\mathrm{dCC}}^T\bm{\delta}_\textbf{q}$. The dCCs increase the probability of finding a new, clash-free conformation close to the desired search direction $\bm{\delta}_\mathbf{q}$ but gives no guarantee as it is still a linearized procedure. As potential clashes can be introduced at different sites given the new perturbation, we perform up to $k$ iterations of adding more clash-avoiding constraints where necessary. This perturbation is denoted NIK$_k$, and the default setting of $k$ is 5. The ultra-high dimensions of molecular conformation spaces allows for a large number of dCCs while still being mobile. Traditional lower-dimensional linkages would suffer almost immediate immobility.

\subsection*{Planner}

The original planner for KGS~\cite{Yao2012} resembles the RRT planner~\cite{lavalle2000rapidly} but has been adapted for high-dimensional conformation spaces where it is not possible to generate a random conformation in the region of interest.
Algorithm~\ref{alg:binnedRRT} outlines the algorithm.
The input is an initial conformation $\mathbf{q}_\text{init}$, an exploration radius $R$, and a number of desired iterations $I$.
Similar to the original RRT, the seed is selected by generating a random conformation, $\mathbf{q}_\text{rand}$.
However, with molecular linkages $\mathbf{q}_\text{rand}$ is typically located far outside the exploration region and is not necessarily a feasible conformation. To prevent the planner from only picking seeds near the border of the exploration region, the seed is selected from a randomly chosen spherical bin centered on $\mathbf{q}_\text{init}$.
Furthermore, to avoid only stepping away from the initial, perturbations do not go toward $\mathbf{q}_\text{rand}$ but rather in a random direction. This procedure is denoted \emph{binned RRT} in the following.

\begin{figure}[bt]
\centering
\includegraphics[width=1.0\textwidth]{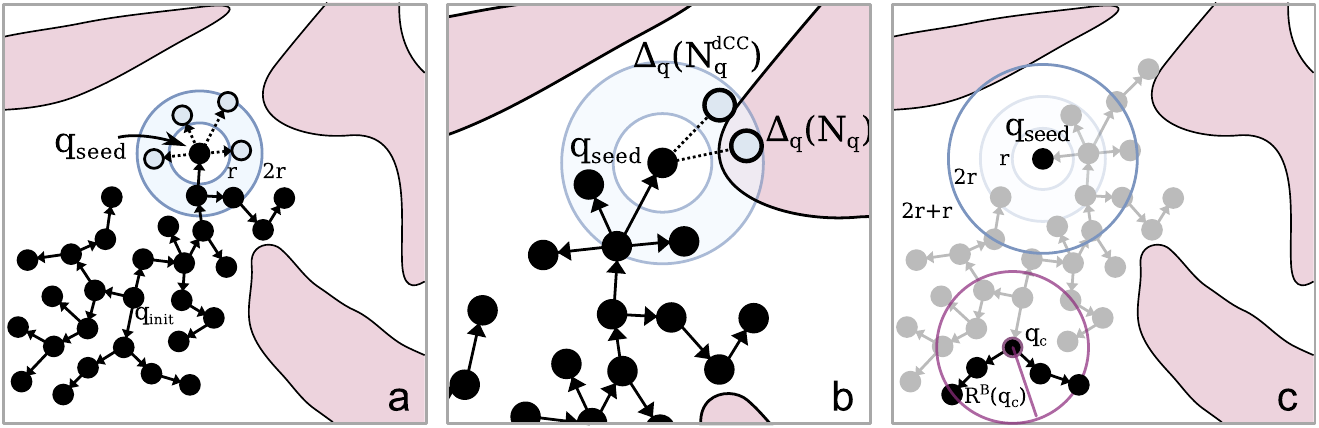}
\caption{High-dimensional Poisson planner. 
a) A random unexplored seed conformation is selected and a Poisson disk with inner (outer) radius $r$ ($2r$) placed around it. 
b) An admissible perturbation within the Poisson disk, $\Delta_q(\mathbf{N}_q)$, is generated by projecting a random perturbation onto the nullspace of $\mathbf{q}_\text{seed}$. 
If a collision occurs we add a dynamic clash constraint, resulting in a new, reduced nullspace $\mathbf{N}_q^\text{dCC}$ with a projection in which the clash is resolved.
c) To avoid oversampling, perturbations are only accepted if they are further than $r$ from any existing conformation. Using a BVH, we maintain the largest distance from $\mathbf{q}_\text{c}$ to any of its descendants, $R^B(\mathbf{q}_c)$. Distance computations between a perturbation of $\mathbf{q}_\text{seed}$ to all descendants of $\mathbf{q}_\text{c}$ can be ignored if $\left|\mathbf{q}_\text{c}\mathbf{q}_\text{seed}\right| > R^B(\mathbf{q}_c) + (2r+r)$, i.e.\ the purple sphere does not collide with the blue sphere. }
\label{fig:explorePanels}
\end{figure}

\floatname{algorithm}{Alg.}
\renewcommand{\algorithmicrequire}{\textbf{Input:}}
\renewcommand{\algorithmicensure}{\textbf{Output:}}
\addtolength{\algorithmicindent}{-2pt}

\begin{figure}[ht]
\noindent
\begin{minipage}[t]{0.46\textwidth}
  \centering
  \begin{algorithm}[H]
  \caption{$\textsc{BinnedRRT}(\mathbf{q}_\textrm{init}, R, \sigma, I)$}
\begin{algorithmic}[H]
\STATE $B \leftarrow $ array of 101 empty bins
\STATE $B[0].\text{add}( \mathbf{q}_\text{init} )$
\FOR{$i=0$ \TO $I$}
\STATE $\mathbf{q}_\text{rand} \leftarrow $ random conformation
\REPEAT
\STATE $b_\text{rand} \leftarrow \textsc{Rand}(0,100)$
\UNTIL{$B[b_\text{rand}] \neq \emptyset$}
\STATE $\mathbf{q}_\text{seed} \leftarrow \argmin_{\mathbf{q}\in B[b_\text{rand}]} \left|\mathbf{q}_\text{rand}\mathbf{q}\right|$
\STATE $\mathbf{q}_\text{new} \leftarrow \textsc{Perturb}(\mathbf{q}_\text{seed}, \sigma)$
\IF{$\textsc{Clash}(\mathbf{q}_\text{new})$}
\STATE $b_\text{new} \leftarrow \lfloor \left|\mathbf{q}_\text{init} \mathbf{q}_\text{new}\right| \cdot \frac{100}{R} \rfloor$
\STATE $B[b_\text{new}].\text{add}( \mathbf{q}_\text{new} )$
\ENDIF
\ENDFOR
\RETURN $\bigcup_{b=0}^{100} B[b]$
\end{algorithmic}
\end{algorithm}
\label{alg:binnedRRT}
\end{minipage}
\hspace{3pt}
\begin{minipage}[t]{0.545\textwidth}
  \centering
  \begin{algorithm}[H]
  \caption{$\textsc{PoissonExplore}(\mathbf{q}_\textrm{init}, r, P)$}
\begin{algorithmic}[H]
\STATE $S_\text{open} \leftarrow \{\mathbf{q}_\text{init} \}$
\STATE $S_\text{closed} \leftarrow \{\}$
\WHILE{$S_\text{open}\neq \emptyset$}
\STATE $\mathbf{q}_\text{seed} \leftarrow S_\text{open}.\text{Pop}()$
\STATE $S' \leftarrow \textsc{BVHCollect}(\mathbf{q}_\text{seed})$
\FOR{$p=0$ \TO $P$}
\STATE $\mathbf{q}_\text{new} \leftarrow \textsc{Perturb}(\mathbf{q}_\text{seed}, \frac{r+2r}{2})$
\IF{$\textsc{Clash}(\mathbf{q}_\text{new}) \land {\forall \mathbf{q}\in S'\ldotp \left|\mathbf{q}_\text{new}\mathbf{q}\right|>r}$}
\STATE $S_\text{open}.\text{add}(\mathbf{q}_\text{new})$
\ENDIF
\ENDFOR
\STATE $S_\text{closed}.\text{add}(\mathbf{q}_\text{seed})$
\ENDWHILE
\RETURN $S_\text{closed}$
\end{algorithmic}
\label{alg:poisson}
\end{algorithm}
\end{minipage}
\caption{Pseudocodes for the RRT-like planner and the Poisson planner. Both algorithms take an initial conformation from which the search is started. Algorithm~\ref{alg:binnedRRT} additionally takes the argument $R$ as the exploration radius around the initial, $\sigma$ the step size of the perturbation, and $I$ the number of iterations. Algorithm 2 takes the argument $r$ as the inner radius of the Poisson disk and $P$ the number of random perturbations that are attempted before closing a seed conformation.}
\end{figure}

The binned RRT planner tends to oversample certain regions, for example near the initial conformation,  while simultaneously limiting fast exploration of unknown territory.
We therefore introduce a multi-query \emph{Poisson planner} (Algorithm~\ref{alg:poisson}), which is based on Poisson disk sampling~\cite{cook1986}. The planner is initialized with a minimum distance $r$ and an initial conformation $\mathbf{q}_\text{init}$ which is added to the open set. At each iteration we randomly select a seed conformation $\mathbf{q}_\text{seed}$ from the set of open conformations (Fig.~\ref{fig:explorePanels}.a) and attempt $P$ perturbations such that the new conformation $\mathbf{q}_\text{new}$ lies within the Poisson disk with inner radius $r$ and outer radius $2r$, i.e., the distance from $\mathbf{q}_\text{new}$ to $\mathbf{q}_\text{seed}$ is between $r$ and $2r$. A perturbation attempt is successful if the resulting conformation is non-clashing and at least a distance $r$ from any existing conformation. Finally, $\mathbf{q}_\text{seed}$ is moved to the set of closed conformations. If there are no more conformations in the open set the procedure ends.

Analyses of existing Poisson sampling algorithms~\cite{bridson2007,dunbar2006} tend to ignore the dimensionality, $n$, of the conformations, so their asymptotic behavior hides an exponential growth in $n$. The crucial step is to quickly find the set of all nearby conformations, $S'$. 
However, the dimension of molecular conformation spaces easily exceeds $n=100$.
Spatial hashing algorithms to identify neighboring samples, proposed in~\cite{bridson2007}, for example, would require checking $3^n\approx 5\cdot 10^{47}$ adjacent bins for neighboring conformations. That is clearly unfeasible.

To address the extremely high dimensionality we designed a bounding volume hierarchy (BVH) algorithm, called \textsc{BVHCollect}, which was inspired by collision detection in protein structures~\cite{lotan2004algorithm,fonseca2012}. 
Each conformation $\mathbf{q}$ is associated with a reference to the seed it was generated from (its parent), all conformations it served as seed to (its children), and the radius $R^B(\mathbf{q})$ of the sphere containing all descendants of $\mathbf{q}$ (Fig.~\ref{fig:explorePanels}.c). 
When a seed is selected, nearby nodes are located by traversing the BVH in a depth-first-manner starting at the root ($\mathbf{q}_\textrm{init}$). A visited node, $\mathbf{q}_\textrm{c}$, and all its descendants can be pruned from the traversal if they are sufficiently far from $\mathbf{q}_\textrm{seed}$ that a collision with a new conformation is impossible:
\begin{equation}
\left|\mathbf{q}_\textrm{c}\mathbf{q}_\textrm{seed}\right| > 2r + r + R^B(\mathbf{q}_\textrm{c})
\end{equation}
Here, $2r$ on the right-hand side is the furthest from $\mathbf{q}_\textrm{seed}$ a new conformation could possibly be placed and $r$ is its empty inner disk radius. After perturbations have been generated, bounding volumes with associated $R^B$-values on the path from $\mathbf{q}_\textrm{seed}$ up to $\mathbf{q}_\textrm{init}$ need to be updated to reflect the new descendants in their subtrees. 

Depending on the shape of conformational space the tree can be arbitrarily unbalanced, so \textsc{BVHCollect} takes linear time worst-case and consequently \textsc{PoissonExplore} takes $\mathcal{O}(N^2)$-time worst-case, where $N$ is the total number of conformations generated. However, there is no $n$ hidden in the exponent, constants associated with traversing the tree are so small that it easily compares with iterating through a list, and in practice large portions of the tree do get pruned.



\section*{Results}

We evaluated the performance of our algorithms on two proteins and an RNA with different sizes and characteristics. Escherichia coli dihydrofolate reductase (PDB ID 3QL3, \emph{DHFR} in the remainder) is a 755 degree of freedom enzyme. We also consider its \emph{M20} (residues 14-25) and \emph{FG} (residues 116-128) loops. The loops share the same underlying molecular graph and initial structure as the full enzyme (PDB ID 3QL3) but while all torsional DOFs are active in DHFR, only loop DOFs are activated when we consider loops. The \emph{Pseudoknot} is the topologically knotted T-arm non-coding RNA of turnip yellow mosaic virus (PDB ID 1a60). Finally, protein \emph{G}$\alpha$\emph{s} represents the alpha subunit of heterotrimeric G-protein (PDB ID 1AZT). Unless otherwise stated, \textsc{PoissonExplore} uses only the NIK$_5$ perturbation and \textsc{BVHCollect} to query for neighboring conformations. In the following we evaluate the  improvements for each of our contributions separately and the last subsection demonstrates a practical application of our methods. 

\subsection*{Acceptance Rate Increases with Dynamic Clash Constraints}

To illustrate that the dCCs used in NIK$_5$ results in higher acceptance rates we ran \textsc{PoissonExplore} with NIK$_0$ (no dCC) and NIK$_5$ perturbations on the M20 loop and our three full systems. Since acceptance rates could be affected by the choice of planner we performed the same tests on a random walk strategy that rejects clashing structures. This strategy resembles frequently used Monte Carlo sampling methods and is therefore labeled MCl.

\noindent
\begin{table}[ht]
\centering
\begin{tabular}{llrrrr}
                               &                  & M20 loop & P-knot & DHFR  & G$\alpha$s \\\hline
\multicolumn{2}{l}{Degrees of freedom}            & 48       & 326    & 755   & 1713       \\\hline
\multirow{2}{*}{\textsc{PoissonExplore}} & Clash rate      & 4\%      & 14\%   & 44\%  & 60\%       \\
                                & Disk reject rate & 6\%      & 1\%    & 1\%   & 1\%        \\\hline
\multirow{2}{*}{\textsc{PoissonExplore} (no dCC)}  & Clash rate      & 88\%     & 81\%   & 88\%  & 79\%       \\
                                & Disk reject rate & 4\%      & 1\%    & 2\%   & 7\%        \\\hline
MCl                         & Clash rate      & 19\%     & 23\%   & 100\% & 100\%      \\\hline
MCl (no dCC)                         & Clash rate      & 91\%     & 89\%   & 100\% & 100\%      \\\hline
\end{tabular}
\vspace{0.25cm}
\caption{Comparison of conformation rejection rates using the \textsc{PoissonExplore} planner or Monte Carlo-like (MCl) sampling using NIK$_5$ and NIK$_0$ (no dCC) perturbations. For \textsc{PoissonExplore} we distinguish between rejections due to clashes (clash rate) and those rejected by failing to meet the Poisson disk criteria for new conformations (disk reject rate).}
\label{tab:rejectRates}
\end{table}

With NIK$_0$, the clash rejection rate of \textsc{PoissonExplore} is independent of system size, and is extremely high, around 80-90\% (Table~\ref{tab:rejectRates}). For all test systems, using NIK$_5$ dramatically reduces conformation rejection rates due to clashes. For the M20 loop specifically the improvement is an order of magnitude. The reductions are still substantial for larger molecular structures, reducing rejections by a factor of two in the case of DHFR, but diminish as the size increases owing to cascading collisions. For larger systems, resolving a collision with NIK$_5$ is more likely to introduce new clashes owing to large, densely packed protein cores. Introducing arbitrarily many collision constraints would rigidify large portions of the molecule, and is not likely result in higher acceptance rates. 


The computational cost of NIK$_5$ is higher than NIK$_0$ as there are more SVD computations to perform. The number of accepted conformations per time-unit for NIK$_5$ is currently half that of NIK$_0$, but a major advantage is that it opens up regions of conformational space that are otherwise difficult to access. Additionally, the collective motions can give insight into molecular mechanisms of conformational change (see Section "\nameref{sec:correlatedMotion}").

Finally, we observe that naive sampling using a random walk strategy (MCl) results in extremely high rejection rates regardless dCC, making a strong case for motion planning-based strategies. 

\subsection*{Bounding Volume Hierarchy Speeds Up Neighbor Search}

To test the efficiency of \textsc{CollectBVH} we recorded the number of distance computations as the  planner explored the native state of the pseudoknot molecule.
\begin{figure}[t]
\centering
\includegraphics[width=0.95\textwidth]{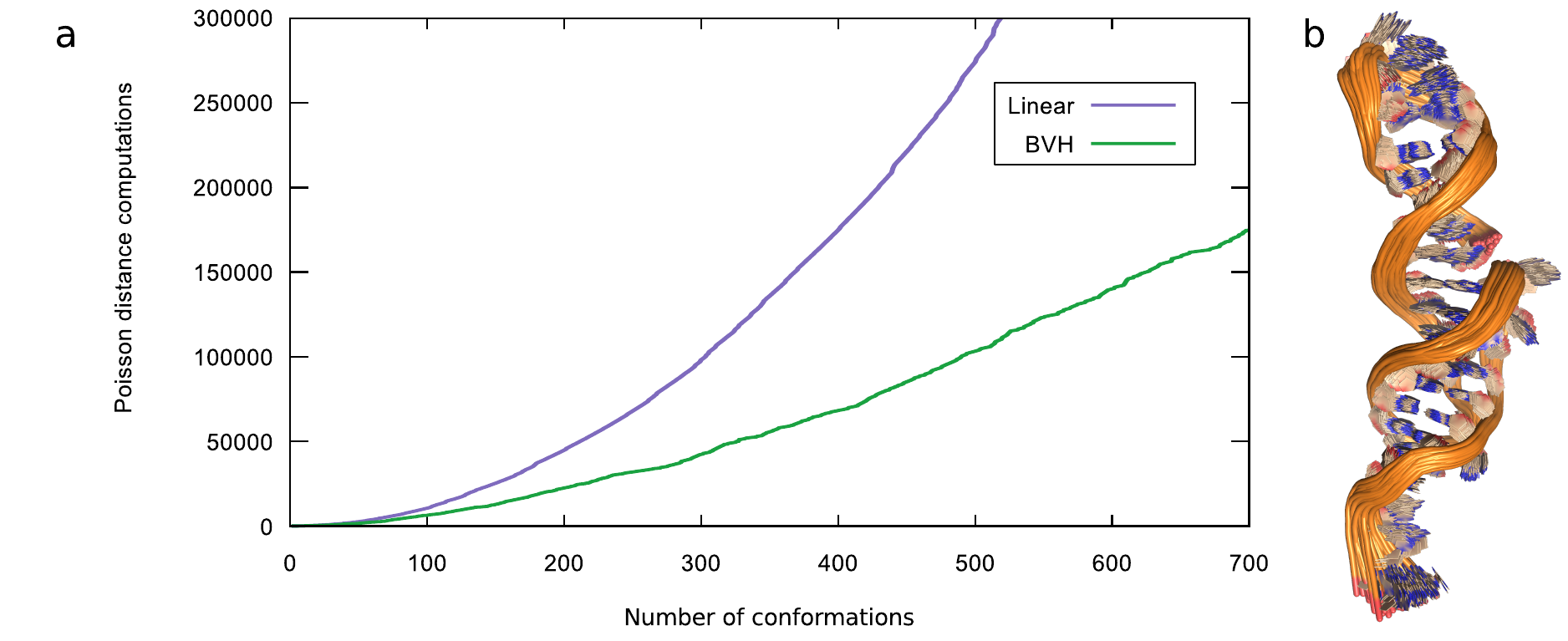}
\caption{Distance computations involved in making a perturbation to the pseudoknot using the BVH versus linearly checking all conformations.}
\label{fig:distanceComputations}
\end{figure}

Fig.~\ref{fig:distanceComputations}.a shows the number of distance computations performed by \textsc{BVHCollect} and the number performed by a linear search through all existing conformations. As expected, the linear search subroutine makes \textsc{PoissonExplore} perform a quadratic number of total distance computations. For the first few hundred conformations, \textsc{BVHCollect} results in about half the number of distance computations but after about 400 conformations the evolution of distance computations takes on a nearly linear trend. This linear-time behavior of \textsc{PoissonExplore} is a best-case performance and can not  be expected for general problems. One explanation might be that the pseudoknot structure is particularly elongated and flexible around the middle (see Fig.~\ref{fig:distanceComputations}.b). \textsc{PoissonExplore} will close conformations near the initial conformation and then explore in separate directions which permits the \textsc{BVHCollect} to efficiently prune large branches in distant regions.

\subsection*{Comparison With RCD+}

To test the sampling quality of \textsc{PoissonExplore}, we generated conformations of the M20 loop in DHFR. This functionally important loop is well-characterized, and adopts three distinct conformations during the catalytic cycle: closed, occluded, and open~\cite{Sawaya1997}. We evaluated the distribution of all-atom RMSD distances to each of these three conformations, and compared our conformational ensembles to the state-of-the-art kinematics-based sampler, random coordinate descent (RCD+, \cite{chys2013}). RCD+ mimics cyclic coordinate descent but selects bonds for optimization randomly, and updates loop conformations by spinor-matrices and geometric filters. 

One immediate observation from this experiment is that the ability of \textsc{PoissonExplore} to fully exhaust the exploration (i.e.\ close all open conformations) depends on finely calibrating the Poisson disk size. If it is too large, $\mathbf{q}_\textrm{init}$ is immediately closed without opening up any new conformations and if it is too small the search will proceed indefinitely. To ensure broad sampling, in this experiment we therefore set $r$ slightly smaller than what would exhaust the search and we set the procedure to terminate after 1,000 conformations. Additionally, before closing a conformation we select all triples of residues that don't have any internal hydrogen bonds and add all possible AIK moves that can be performed on the C$_\alpha$ atoms. 

\begin{figure}[t]
\centering
\includegraphics[width=0.7\textwidth]{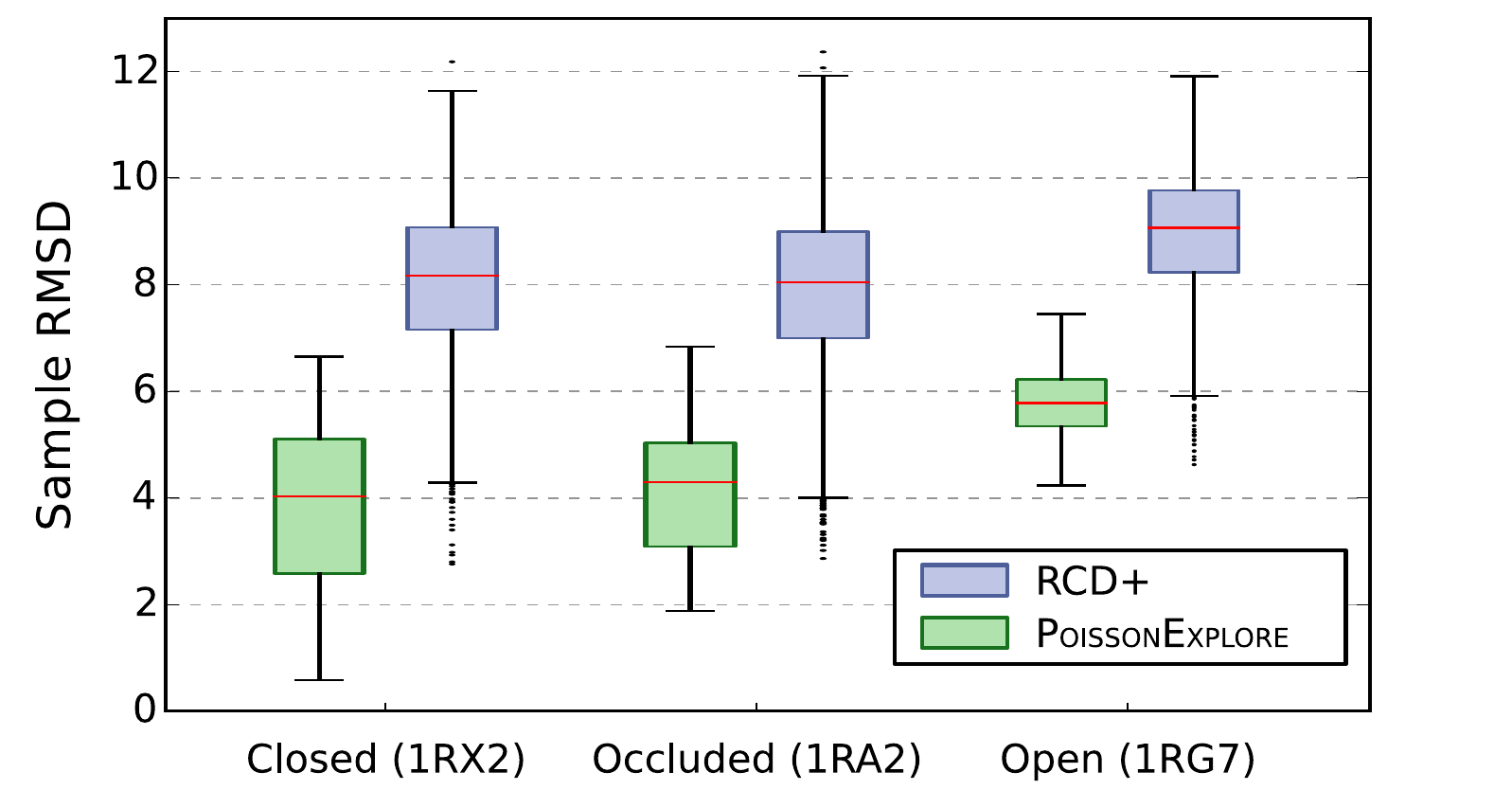}
\caption{Accuracy of loop sampling.
Using \textsc{PoissonExplore} with AIK and NIK$_5$ as well as RCD+ we sampled 1,000 conformations of the M20 loop (residue 14-25) and measured their distances to 3 loop conformations known to represent a closed, an occluded and an open conformation. \textsc{PoissonExplore} was initialized near the closed conformation while RCD+ completely rebuilds loops in each iteration. The middle 50th percentile of each distribution is represented as green and blue boxes.}
\label{fig:RCD}
\end{figure}

Both methods sample a broad ensemble of states (Fig.~\ref{fig:RCD}). For all three cases, however, \textsc{PoissonExplore} samples closer to the experimentally observed conformation than RCD+.
Importantly, for the closed and occluded states, the middle 50th percentile of \textsc{PoissonExplore} conformations extend to a broader RMSD range than the comparable RCD+ ensemble. We anticipate this behavior as the Poisson disks result in a dense set of approximately equidistant conformations. However, RCD+ conformations extend to an {\em overall} larger RMSD range for all three states. This could be ascribed to the fact that \textsc{PoissonExplore} diffuses more slowly through conformation space than a random sampler like RCD+. A larger number of conformations could possibly help this out. Additionally, the \textsc{PoissonExplore} conformations may be more constrained by collision-avoidance than RCD+: Among the 1,000 RCD+ loop conformations, 120 have collisions even when van der Waals radii are scaled by 75\%, i.e., they would have been rejected by our method.

\subsection*{Correlated Motions in DHFR Active Site}
\label{sec:correlatedMotion}

Next, we examine if the kinematic cycles defined by native and non-native contacts propagate collective motions in proteins. The functionally important FG loop (residues 116-128) in ecDHFR connects the F and G $\beta$-strands. A 'distal' amino acid mutation G121V in this loop reduces catalytic activity of the enzyme 200-fold, despite the fact that the residue is nearly 15\AA{} from the active site. Nuclear Magnetic Resonance (NMR) spectroscopy data furthermore suggests that the FG loop and the M20 loop are dynamically linked~\cite{Boehr2013,VandenBedem2013a}. There is only one hydrogen bond between the loops (G15/O to D122/H). We generated 1,000 conformations for the FG and M20 loops simultaneously, by proposing trial moves $\mathbf{\Delta}_{q}$ for the rotateable bonds in both loops and projecting $\mathbf{\Delta}_{q}$ onto the nullspace of the entire molecule. This drives motion mainly in the loop regions, but permits the rest of the molecule to adapt to these motions.

We analyzed collective motions of the two loops by computing the correlations between the positions of their C$_\alpha$ atoms over all conformations. The correlation between atoms $i$ and $j$ is characterized by the quantity~\cite{Kamberaj2009}
\begin{equation}
C_{i,j} = \frac{\left< \Delta\mathbf{p}_i(q)\cdot \Delta\mathbf{p}_j(q) \right>_q}{  \sqrt{\left<\left|\Delta\mathbf{p}_i(q)\right|^2\right>_q \cdot \left<\left|\Delta\mathbf{p}_j(q)\right|^2\right>_q}  }
\end{equation}
where $\left<\right>_q$ denotes the average over all conformations, $\mathbf{p}_i(q)$ is the position of atom $i$ in conformation $q$, and $\Delta\mathbf{p}_i(q) = \mathbf{p}_i(q) - \left< \mathbf{p}_i(q)\right>_q$. Note that the normalization term in the denominator results in downscaling of $C_{i,j}$ if there is little motion. 

\begin{figure}[ht]
\centering
\includegraphics[width=1.0\textwidth]{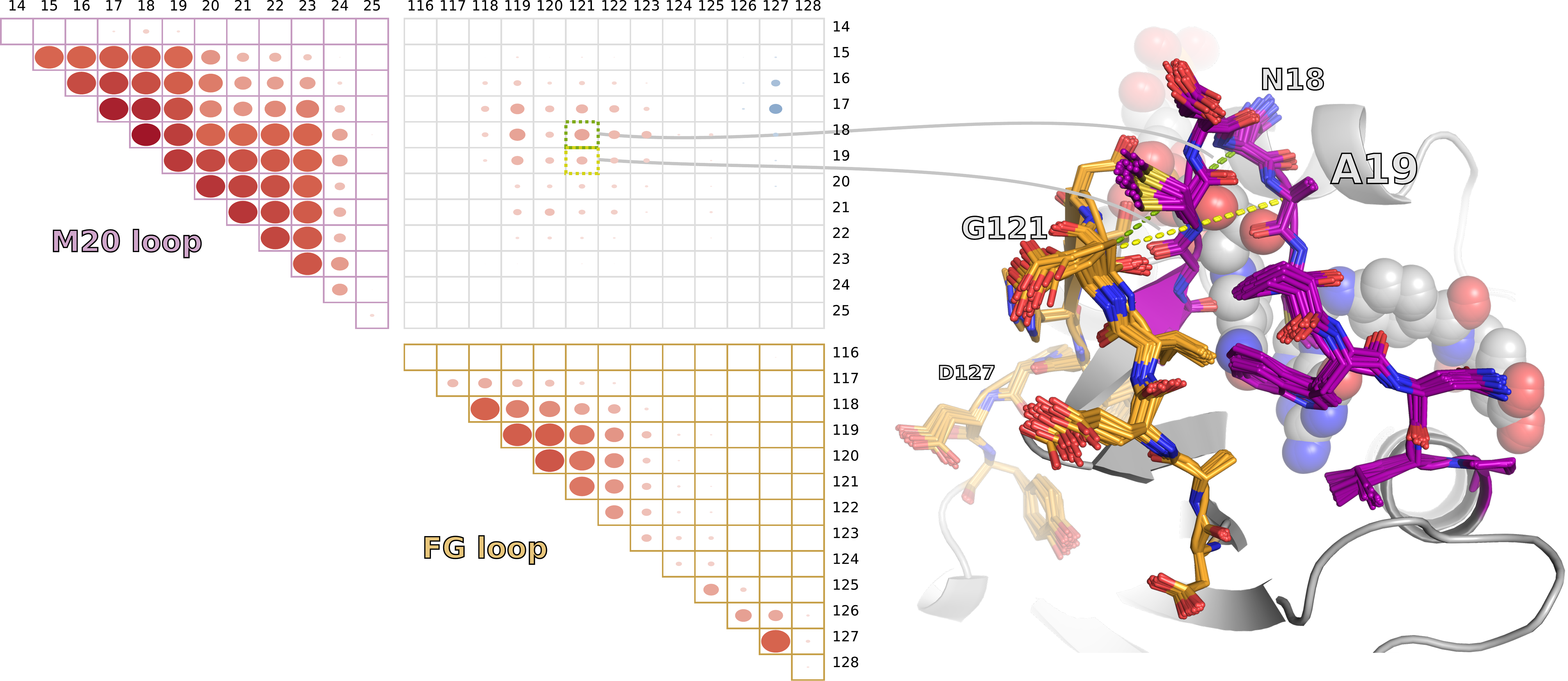}
\caption{\label{fig:DHFRcorr}Correlated loop motions in DHFR.
Both the M20 loop and the FG loop of DHFR were sampled using the Poisson planner with dCC and the correlation between C$_\alpha$ atoms in each loop plotted. 
Red colors indicate high correlation ($C_{i,j}$) and blue colors anti-correlation (negative $C_{i,j}$). 
Inter-loop pairs with particularly high correlation are highlighted in the structural models on the right. 
}
\end{figure}

We observed a higher degree of self-correlation within the M20 loop than in the FG loop (Fig.~\ref{fig:DHFRcorr}). The first half of the M20 loop backbone shows slightly reduced correlations with the second half, suggesting somewhat independent motions. Proline P21 and tryptophan W22, a more rigid and a bulky side-chain, likely reduce the feasible region of conformational space for the second half. Correlations between the two halves of the FG loop are highly reduced, with the start of the loop showing more motion. The reduced self-correlations of residues reflect smaller motions overall. 

Interestingly, the highest inter-loop correlations are seen between residue pairs that are not in direct contact, but communicate through steric collisions and the hydrogen bond network. G121 of the FG loop clearly stands out, collectively moving with N18 and A19 of the M20 loop. A19 is hydrogen bonded to M16, which sterically interacts with G121. Thus, both native and non-native contacts propagate these collective motions.

Residues D127 and E17 stand out by their anti-correlated motions. We propose that these residues are the results of a somewhat asymmetric hinge motion around the last half of the FG loop driven by interactions with the M20 loop. The lack of correlations for residues 124-126 supports this claim; when E17 on top of the M20 loop moves upwards, the hinge turns and D127 moves downwards, and vice versa. 

\section*{Conclusion}

The ultra-high dimensions of molecular conformation spaces require efficient methods to broadly and uniformly sample conformations and relate these conformational sets to biological function. We combine different molecular perturbations from analytical inverse kinematics solutions, nullspace projections and our new, dynamic clash-avoiding constraint strategy, to approach a comprehensive representation of the underlying constraint variety. Our four examples consistently illustrate that this strategy increases acceptance rates in highly rugged search spaces. Coupled to an adapted Poisson disk inspired motion planner with bounding volume hierarchy (\textsc{PoissonExplore}), our algorithm efficiently samples distinct conformations in the ultra-high conformation space close to experimentally known functional states. Compared to RCD+ on DHFR's M20 loop, we obtain broad conformational ensembles around the known closed, occluded, and open state.
Finally, we analyzed how functional motions in DHFR are coupled through native and non-native contacts between the M20 and the FG-loop. Surprisingly, correlation analysis reveals coordinated motions in both loops linked to residues that are not direct neighbors. This reveals how motions encoded by our clash-avoiding constraints are propagated over long distances, revealing underlying functional mechanisms in the molecule. Our method is generally applicable to ultra-high dimensional problems in robotics and related research fields, subject to holonomic constraints and (self-) collisions. Software and examples are available at \url{https://simtk.org/projects/kgs/}.

\bibliographystyle{plain}
\bibliography{bibliography}

\begin{thebibliography}{10}

\bibitem{Al-Bluwi2012b}
Ibrahim Al-Bluwi, Thierry Sim{\'{e}}on, and Juan Cort{\'{e}}s.
\newblock {Motion planning algorithms for molecular simulations: A survey}.
\newblock {\em Comput Sci Rev}, 6(4):125--143, jul 2012.

\bibitem{al2013modeling}
Ibrahim Al-Bluwi, Marc Vaisset, Thierry Sim{\'e}on, and Juan Cort{\'e}s.
\newblock Modeling protein conformational transitions by a combination of
  coarse-grained normal mode analysis and robotics-inspired methods.
\newblock {\em BMC Struct Biol}, 13(1):1, 2013.

\bibitem{Boehr2013}
David~D. Boehr, Jason~R. Schnell, Dan McElheny, Sung-Hun Bae, Brendan~M.
  Duggan, Stephen~J. Benkovic, H.~Jane Dyson, and Peter~E. Wright.
\newblock A distal mutation perturbs dynamic amino acid networks in
  dihydrofolate reductase.
\newblock {\em Biochemistry}, 52(27):4605--4619, 2013.

\bibitem{bridson2007}
Robert Bridson.
\newblock Fast poisson disk sampling in arbitrary dimensions.
\newblock In {\em SIGGRAPH sketches}, page~22, 2007.

\bibitem{budday2016frustration}
Dominik Budday, Rasmus Fonseca, Sigrid Leyendecker, and Henry van~den Bedem.
\newblock Frustration-guided motion planning reveals conformational transitions
  in proteins.
\newblock Submitted.

\bibitem{Budday2015a}
Dominik Budday, Sigrid Leyendecker, and Henry van~den Bedem.
\newblock {Geometric analysis characterizes molecular rigidity in generic and
  non-generic protein configurations}.
\newblock {\em J Mech Phys Solids}, 83:36--47, 2015.

\bibitem{Burdick1989}
J~W Burdick.
\newblock {On the inverse kinematics of redundant manipulators:
  characterization of the self-motion manifolds}.
\newblock In {\em Proc IEEE Int Conf Robot Autom}, pages 264--270. IEEE Comput.
  Soc. Press, 1989.

\bibitem{chys2013}
Pieter Chys and Pablo Chac{\'o}n.
\newblock Random coordinate descent with spinor-matrices and geometric filters
  for efficient loop closure.
\newblock {\em J Chem Theory Comput}, 9(3):1821--1829, 2013.

\bibitem{cook1986}
Robert~L Cook.
\newblock Stochastic sampling in computer graphics.
\newblock {\em ACM Transactions on Graphics (TOG)}, 5(1):51--72, 1986.

\bibitem{cortes2005path}
Juan Cort{\'e}s, Thierry Sim{\'e}on, V~Ruiz De~Angulo, David Guieysse, Magali
  Remaud-Sim{\'e}on, and Vinh Tran.
\newblock A path planning approach for computing large-amplitude motions of
  flexible molecules.
\newblock {\em Bioinformatics}, 21(suppl 1):i116--i125, 2005.

\bibitem{coutsias2006}
Evangelos~A Coutsias, Chaok Seok, Michael~J Wester, and Ken~A Dill.
\newblock Resultants and loop closure.
\newblock {\em International Journal of Quantum Chemistry}, 106(1):176--189,
  2006.

\bibitem{angulo2005}
V.R. de~Angulo, Thierry Sim{\'{e}}on, and Juan Cort{\'{e}}s.
\newblock Biocd: An efficient algorithm for self-collision and distance
  computation between highly articulated molecular models.
\newblock In {\em Proceedings of Robotics: Science and Systems}, 2005.

\bibitem{Dror2012-ai}
Ron~O Dror, Robert~M Dirks, J~P Grossman, Huafeng Xu, and David~E Shaw.
\newblock Biomolecular simulation: a computational microscope for molecular
  biology.
\newblock {\em Annu Rev Biophys}, 41:429--452, January 2012.

\bibitem{dunbar2006}
Daniel Dunbar and Greg Humphreys.
\newblock A spatial data structure for fast poisson-disk sample generation.
\newblock {\em ACM Transactions on Graphics (TOG)}, 25(3):503--508, 2006.

\bibitem{Fonseca02092014}
Rasmus Fonseca, Dimitar~V. Pachov, Julie Bernauer, and Henry van~den Bedem.
\newblock Characterizing rna ensembles from nmr data with kinematic models.
\newblock {\em Nucl Acids Res}, 42(15):9562--9572, 2014.

\bibitem{fonseca2012}
Rasmus Fonseca and Pawel Winter.
\newblock Bounding volumes for proteins: a comparative study.
\newblock {\em Journal of Computational Biology}, 19(10):1203--1213, 2012.

\bibitem{golub2012matrix}
Gene~H Golub and Charles~F Van~Loan.
\newblock {\em Matrix computations}, volume~3.
\newblock JHU Press, 2012.

\bibitem{halperin1994}
Dan Halperin and Mark~H Overmars.
\newblock Spheres, molecules, and hidden surface removal.
\newblock In {\em Proceedings of the tenth annual symposium on Computational
  geometry}, pages 113--122. ACM, 1994.

\bibitem{haspel2010tracing}
Nurit Haspel, Mark Moll, Matthew~L Baker, Wah Chiu, and Lydia~E Kavraki.
\newblock Tracing conformational changes in proteins.
\newblock {\em BMC Struct Biol}, 10(Suppl 1):S1, 2010.

\bibitem{Kamberaj2009}
Hiqmet Kamberaj and Arjan van~der Vaart.
\newblock Correlated motions and interactions at the onset of the dna-induced
  partial unfolding of ets-1.
\newblock {\em Biophys J}, 96(4):1307--1317, 2009.

\bibitem{kavraki1996probabilistic}
Lydia~E Kavraki, Petr {\v{S}}vestka, Jean-Claude Latombe, and Mark~H Overmars.
\newblock Probabilistic roadmaps for path planning in high-dimensional
  configuration spaces.
\newblock {\em Robotics and Automation, IEEE Transactions on}, 12(4):566--580,
  1996.

\bibitem{kirillova2008nma}
Svetlana Kirillova, Juan Cort{\'e}s, Alin Stefaniu, and Thierry Sim{\'e}on.
\newblock An nma-guided path planning approach for computing large-amplitude
  conformational changes in proteins.
\newblock {\em Proteins: Struct, Funct, Bioinf}, 70(1):131--143, 2008.

\bibitem{lagae2008comparison}
Ares Lagae and Philip Dutr{\'e}.
\newblock A comparison of methods for generating poisson disk distributions.
\newblock In {\em Computer Graphics Forum}, volume~27, pages 114--129. Wiley
  Online Library, 2008.

\bibitem{lavalle2000rapidly}
Steven~M LaValle and James~J Kuffner~Jr.
\newblock Rapidly-exploring random trees: Progress and prospects.
\newblock 2000.

\bibitem{lotan2004algorithm}
Itay Lotan, Fabian Schwarzer, Dan Halperin, and Jean-Claude Latombe.
\newblock Algorithm and data structures for efficient energy maintenance during
  monte carlo simulation of proteins.
\newblock {\em J Comput Biol}, 11(5):902--932, 2004.

\bibitem{Noe2006}
F~No\'e, Dieter Krachtus, Jeremey~C. Smith, and S.~Fischer.
\newblock {Transition Networks: Computational Methods for the Comprehensive
  Analysis of Complex Rearrangements in Proteins}.
\newblock {\em J Chem Theory Comput}, 2:840--857, 2006.

\bibitem{Oliveberg2005}
Mikael Oliveberg and Peter~G Wolynes.
\newblock {The experimental survey of protein-folding energy landscapes.}
\newblock {\em Q Rev Biophys}, 38:245--288, 2005.

\bibitem{park2014poisson}
Chonhyon Park, Jia Pan, and Dinesh Manocha.
\newblock {Poisson-RRT}.
\newblock In {\em 2014 IEEE International Conference on Robotics and Automation
  (ICRA)}, pages 4667--4673. IEEE, 2014.

\bibitem{Petric2013}
Tadej Petri{\v{c}} and Leon {\v{Z}}lajpah.
\newblock {Smooth continuous transition between tasks on a kinematic control
  level: Obstacle avoidance as a control problem}.
\newblock {\em Rob Aut. Syst}, 61(9):948--959, 2013.

\bibitem{Raveh2009}
B.~Raveh, A.~Enosh, O.~Schueler-Furman, and D.~Halperin.
\newblock Rapid sampling of molecular motions with prior information
  constraints.
\newblock {\em {PLoS} Comput Biol}, 5:e1000295, 2009.

\bibitem{Sawaya1997}
M~R Sawaya and J~Kraut.
\newblock {Loop and subdomain movements in the mechanism of Escherichia coli
  dihydrofolate reductase: crystallographic evidence.}
\newblock {\em Biochemistry}, 36(3):586--603, jan 1997.

\bibitem{Sugiura2007}
Hisashi Sugiura, Michael Gienger, Herbert Janssen, and Christian Goerick.
\newblock {Real-time collision avoidance with whole body motion control for
  humanoid robots}.
\newblock In {\em IEEE Int. Conf. Intell. Robot. Syst.}, pages 2053--2058,
  2007.

\bibitem{thomas2007simulating}
Shawna Thomas, Xinyu Tang, Lydia Tapia, and Nancy~M Amato.
\newblock Simulating protein motions with rigidity analysis.
\newblock {\em J Comput Biol}, 14(6):839--855, 2007.

\bibitem{VandenBedem2013a}
Henry van~den Bedem, Gira Bhabha, Kun Yang, Peter~E Wright, and James~S Fraser.
\newblock {Automated identification of functional dynamic contact networks from
  X-ray crystallography.}
\newblock {\em Nat Meth}, 10(9):896--902, 2013.

\bibitem{VandenBedem2015}
Henry van~den Bedem and James~S Fraser.
\newblock {Integrative, dynamic structural biology at atomic resolution--it's
  about time}.
\newblock {\em Nat Meth}, 12(4):307--318, 2015.

\bibitem{VandenBedem2005}
Henry van~den Bedem, Itay Lotan, Jean~Claude Latombe, and Ashley~M Deacon.
\newblock {Real-space protein-model completion: an inverse-kinematics
  approach.}
\newblock {\em Acta Cryst}, D61:2--13, 2005.

\bibitem{Yao2012}
Peggy Yao, Liangjun Zhang, and Jean-Claude Latombe.
\newblock Sampling-based exploration of folded state of a protein under
  kinematic and geometric constraints.
\newblock {\em Proteins: Struct, Funct, Bioinf}, 80(1):25--43, 2012.

\end{thebibliography}
\end{document}